\documentclass[11pt]{article}
\usepackage{acl}

\usepackage{times}
\usepackage{latexsym}
\usepackage[T1]{fontenc}
\usepackage[utf8]{inputenc}
\usepackage{microtype}
\usepackage{amsmath,amssymb,amsthm}
\usepackage{booktabs}
\usepackage{multirow}
\usepackage{float}
\usepackage{graphicx}
\usepackage{xcolor}
\usepackage{tikz}
\usepackage{authblk}
\usetikzlibrary{positioning,arrows.meta,calc,fit}

\newif\ifshowtodos \showtodostrue

\newcommand{\cA}{c_{\mathrm{A}}}
\newcommand{\cD}{c_{\mathrm{D}}}
\newcommand{\LN}{\mathrm{LN}}
\newcommand{\FFN}{\mathrm{FFN}}
\newcommand{\up}{\!\uparrow}

\newcommand*{\affaddr}[1]{#1} % No op here. Customize it for different styles.
\newcommand*{\affmark}[1][*]{\textsuperscript{#1}}
\newcommand*{\email}[1]{\texttt{#1}}

\title{WNet: Discrete Wavelets Transform for Efficient Token Mixing}

\author{%
\bf{Rana Salama \affmark[1,]\affmark[2], Abdou Youssef \affmark [1] and Mona Diab \affmark[3]}\\
\affaddr{\affmark[1]George Washington University, Washington DC, USA }\\
\affaddr{\affmark[2]Faculty of Computers and Artificial Intelligence, Cairo University, Egypt}\\
\affaddr{\affmark[3]Carnegie Mellon University, Pittsburgh, Pennsylvania, USA}\\
\email{raref@gwu.edu, ayoussef@gwu.edu , mdiab@andrew.cmu.edu}\\\
}

\begin{document}
\maketitle
% ===========================================================================
\begin{abstract}
In a Transformer, token mixing is the step that lets each token draw information from other             tokens, and it dominates the cost of encoding long sequences. Self-attention does this mixing very well: every token weighs every other token by content, which gives strong contextual modeling. That all-pairs comparison is also why its cost grows quadratically with sequence length. We introduce WNet, a Transformer encoder that replaces self-attention with token mixing based on the discrete wavelet transform (DWT). Three attention-free mixers recombine the scales: by linear fusion, by learned gating, or by letting each token choose its scales. A hybrid adds self-attention in the last layer only. A receptive-field analysis shows that wavelet mixers built from two-tap filters, such as Haar, never relate tokens outside fixed blocks, however deep the network, even when the filters are learned. Longer filters reach the whole sequence within two layers. We pre-train every model with masked language modeling on a fixed-token subset of C4 and fine-tune on GLUE, using one controlled setup with size-matched BERT and FNet baselines and a control that cannot mix tokens. The token-gated mixer trains as fast as attention at 256 tokens and 2.7$\times$ faster at 4,096. 
%\todo{Main quality result of the full run: WNet-Dyn and WNet-Hybrid vs.\ BERT on MNLI and pre-training loss.}
\end{abstract}

% ===========================================================================
\section{Introduction}

Self-attention \citep{vaswani2017attention} lets every token weigh every other token by content, which makes it a powerful model of context. It is also expensive: its cost grows quadratically with the sequence length $S$. FNet \citep{leethorp2022fnet} showed that much of what attention contributes on standard benchmarks can be recovered with far less. It replaces each attention sublayer with a fixed Fourier transform and retains 92--97\% of BERT's GLUE accuracy \citep{devlin2019bert,wang2019glue} while training faster.

A Fourier coefficient, however, weighs all tokens equally, so the transform has no notion of locality. Language is organised hierarchically: morphology and syntax act over short spans and discourse over long ones. DWT \citep{mallat1989theory,daubechies1992ten} matches this organisation. It splits a signal into a coarse summary and local details at successively coarser scales, in $\mathcal{O}(S)$ time.

We introduce \textbf{WNet}, a Transformer encoder in which every attention sublayer is replaced by a wavelet token mixer (Fig.~\ref{fig:arch}).
\begin{itemize}
\setlength\itemsep{0pt}
\item \textbf{Decomposition.} The mixer decomposes each channel of the hidden states into a few scales. Each scale is kept at the full length of the sequence, so that every token has a value at every scale. The scales add up exactly to the input, and shifting the input shifts every scale.
\item \textbf{Recombination.} A light learned layer recombines the scales: by linear fusion (WNet-MS), by learned gating (WNet-G), or by letting each token choose its scales (WNet-Dyn).
\item \textbf{Hybrid.} WNet-Hybrid adds self-attention in the last layer only.
\end{itemize}

A wavelet mixer is local by design, so the central question is how far information can travel through a stack of such layers. We answer it with an exact receptive-field analysis. Mixers built from two-tap filters, which include every Haar wavelet design and recently proposed learnable Haar mixing \citep{kiruluta2025lmwt}, only ever relate tokens inside fixed blocks, however deep the network. Longer filters remove the blocks and reach the whole sequence within two layers.

Our contributions are:
\begin{itemize}
\setlength\itemsep{0pt}
\item \textbf{WNet}, a family of linear-time wavelet token mixers built on a full-resolution multiscale decomposition.
\item A \textbf{receptive-field analysis} of wavelet mixers. It shows that two-tap designs are confined to fixed blocks at any depth, even with learned filters, and that the standard, downsampled DWT represents a phrase differently depending on where it starts.
\item A \textbf{controlled comparison} in which every model shares data, depth, width, optimiser and budget. The comparison includes size-matched BERT and FNet, a re-implementation of learnable Haar mixing and a control that cannot mix tokens. We also report \textbf{speed} up to 4,096 tokens.
\end{itemize}

% ===========================================================================
\section{Related Work}

\paragraph{Efficient attention.} Several families of methods reduce the cost of attention while keeping its content-dependent weights \citep{tay2022survey}: sparse patterns \citep{beltagy2020longformer,zaheer2020bigbird}, low-rank projections \citep{wang2020linformer} and kernel approximations \citep{choromanski2021performer}. Others shorten the sequence before attention with pooling or a cosine transform \citep{dai2020funnel,he2023fouriertransformer,scribano2023dctformer}. These methods approximate attention, whereas WNet replaces it.

\paragraph{Attention-free mixing.} MLP-Mixer \citep{tolstikhin2021mlpmixer} and gMLP \citep{liu2021gmlp} mix tokens with learned dense or gated projections. FNet \citep{leethorp2022fnet} uses a fixed Fourier transform, and GFNet \citep{rao2021gfnet} and AFNO \citep{guibas2022afno} learn filters in the Fourier domain. Other mixers approach attention quality on long sequences: long convolutions with multiplicative gating \citep{poli2023hyena}, structured matrices \citep{fu2023monarch} and state-space models \citep{gu2022s4,gu2024mamba}. Recent spectral mixers add content-adaptive or causal mixing \citep{feinashley2025spectre,gan2026caracal}. WNet-Dyn's gating follows gated linear units \citep{dauphin2017glu}, but the mixing itself remains a fixed wavelet transform.

\paragraph{Wavelets in neural networks.} Several vision models use wavelets: WaveMix \citep{jeevan2022wavemix} mixes image tokens with a 2D DWT, Wave-ViT \citep{yao2022wavevit} downsamples keys and values with wavelets, and SVT \citep{patro2023svt} uses complex wavelets. For sequences, WavSpA \citep{zhuang2024wavspa} computes attention in a wavelet basis and Wavelet GPT \citep{verma2024waveletgpt} adds multi-scale structure to an attention model; both keep attention. Closest to our work is LMWT \citep{kiruluta2025lmwt}. It replaces attention with a five-level Haar decomposition whose two-tap weights are learned per channel, and reaches 27.2 BLEU on WMT16 En--De against 27.8 for attention. %Our analysis shows that this design only relates tokens inside fixed blocks of 32 (\S\ref{sec:analysis}). We re-implement it under our protocol.

% ===========================================================================
\section{Discrete Wavelet Transform}
\label{sec:dwt}

DWT is a multiresolution analysis framework that hierarchically decomposes a signal $f(t)$ through recursive filtering and downsampling. At each level $j$, $f(t)$ is convolved with low-pass and high-pass filters to produce \textit{Approximation coefficients} ($cA_j$), representing low varying components, and \textit{Detail coefficients} ($cD_j$), capturing fine-scale or high variations. These two coefficients togeeather recover the input exactly. Repeating the step on the approximation gives coarser and coarser summaries.

In general, level $j$ filters the previous approximation with a low-pass filter $h$ and a high-pass filter $g$ and keeps every second value:
\begin{equation}
\begin{aligned}
{\cA}_j[k]&=\textstyle\sum_n h[n{-}2k]\,{\cA}_{j-1}[n],\\
{\cD}_j[k]&=\textstyle\sum_n g[n{-}2k]\,{\cA}_{j-1}[n],
\end{aligned}
\label{eq:dwt}
\end{equation}
with ${\cA}_0=x$. After $J$ levels, the input is represented by one approximation of length $S/2^J$ and details of lengths $S/2^J,\dots,S/2$. Haar filters have two taps. Daubechies filters such as db10 (20 taps) are longer and smoother, so each coefficient summarises a wider stretch of tokens. A Fourier coefficient depends on the whole sequence, whereas a wavelet coefficient describes one region at one scale (App.~\ref{app:dwt}).

% ===========================================================================
\section{WNet}
\label{sec:method}

\begin{figure*}[t]
\centering
\resizebox{\textwidth}{!}{%
\begin{tikzpicture}[font=\footnotesize,
  blk/.style={draw, rounded corners=2pt, minimum height=6.5mm, inner sep=3pt, align=center, fill=#1}, blk/.default=gray!8,
  arr/.style={-{Stealth[length=1.6mm]}, thick, gray!70!black},
  sig/.style={inner sep=1.5pt}]
% ---- (a) block
\node[anchor=west, font=\footnotesize\bfseries] at (-0.3,1.45) {(a) WNet block};
\node (x) at (0,0) {$X$};
\node (ln) [blk, right=5mm of x] {LN};
\node (M) [blk=orange!22, right=5mm of ln] {wavelet mixer $\mathcal{M}$\\{\scriptsize decompose $\to$ recombine}};
\node (p1) [circle, draw, inner sep=0.6pt, right=5mm of M] {$+$};
\node (ln2) [blk, right=5mm of p1] {LN};
\node (ffn) [blk=blue!12, right=5mm of ln2] {FFN};
\node (p2) [circle, draw, inner sep=0.6pt, right=5mm of ffn] {$+$};
\node (h) [right=5mm of p2] {$H$};
\foreach \u/\v in {x/ln,ln/M,M/p1,p1/ln2,ln2/ffn,ffn/p2,p2/h} \draw[arr] (\u) -- (\v);
\draw[arr] ($(x.east)+(1.2mm,0)$) -- ++(0,10mm) -| (p1.north);
\draw[arr] ($(p1.east)+(1.2mm,0)$) -- ++(0,10mm) -| (p2.north);
\node[draw, dashed, gray, rounded corners, fit=(ln)(p2), inner sep=3.5mm, label={[gray,font=\scriptsize]north east:$\times 4$}] {};
% ---- (b) decomposition
\begin{scope}[yshift=-22mm]
\node[anchor=west, font=\footnotesize\bfseries] at (-0.3,1.0) {(b) Full-resolution decomposition ($J{=}3$): smooth, then subtract};
\node[sig] (a0) at (0,0) {$A_0{=}U$};
\node[blk=orange!12] (s1) at (19mm,0) {$s_{\up 1}*$};
\node[sig] (a1) at (36mm,0) {$A_1$};
\node[blk=orange!12] (s2) at (53mm,0) {$s_{\up 2}*$};
\node[sig] (a2) at (70mm,0) {$A_2$};
\node[blk=orange!12] (s3) at (87mm,0) {$s_{\up 4}*$};
\node[sig] (a3) at (104mm,0) {$A_3$};
\foreach \u/\v in {a0/s1,s1/a1,a1/s2,s2/a2,a2/s3,s3/a3} \draw[arr] (\u) -- (\v);
\foreach \p/\sm/\c/\n in {a0/s1/a1/1, a1/s2/a2/2, a2/s3/a3/3} {
  \node[circle, draw, inner sep=0pt, minimum size=3.2mm] (m\n) at ($(\sm)+(0,-9mm)$) {};
  \draw (m\n.west) ++(0.7mm,0) -- ++(1.8mm,0);
  \draw[arr] (\p.south) |- (m\n.west) node[pos=0.75, above, font=\scriptsize] {$+$};
  \draw[arr] (\c.south) |- (m\n.east) node[pos=0.75, above, font=\scriptsize] {$-$};
  \node[sig] (d\n) at ($(m\n)+(0,-7mm)$) {$D_\n$};
  \draw[arr] (m\n) -- (d\n);
}
\node[draw, rounded corners=2pt, fill=gray!6, align=left, anchor=west, inner sep=3pt] at (118mm,-8mm)
  {bands $B=(A_3,D_3,D_2,D_1)$\\[1pt]
   $\bullet$ each of length $S$\\
   $\bullet$ $A_3{+}D_3{+}D_2{+}D_1=U$\\
   $\bullet$ shift $U$ $\Rightarrow$ every band shifts\\
   $\bullet$ one convolution per level};
\end{scope}
% ---- (c) recombinations
\begin{scope}[yshift=-52mm]
\node[anchor=west, font=\footnotesize\bfseries] at (-0.3,0.45) {(c) Recombinations};
\foreach \x/\fl/\ta/\tb in {
  0/blue!8/{WNet-MS: fusion}/{$[A_3\Vert D_3\Vert D_2\Vert D_1]\,W$},
  46/blue!12/{WNet-G: learned gating}/{$\mathrm{GELU}(\textstyle\sum_j\alpha_j B_j)\odot\sum_j\beta_j B_j$},
  92/blue!20/{WNet-Dyn: per-token scales}/{$\big(\textstyle\sum_j\sigma_j(X)\odot B_j\big)\odot XW_v$},
  138/green!12/{WNet-Hybrid}/{WNet-Dyn, with self-attention\\in the last layer}} {
  \node[draw, rounded corners=2pt, fill=\fl, anchor=north west, minimum width=43mm, text width=42mm, minimum height=11mm, align=center, inner sep=2pt]
       at (\x mm-2mm,0) {\textbf{\ta}\\ \tb};
}
\end{scope}
\end{tikzpicture}}
\caption{WNet. (a) Each attention sublayer of a pre-LN encoder is replaced by a wavelet mixer. (b) The mixer decomposes its input $U$ along the sequence with a smoothing filter $s$ that is dilated at every level; each detail is what one more smoothing step removes. (c) A learned layer recombines the bands.}
\label{fig:arch}
\end{figure*}
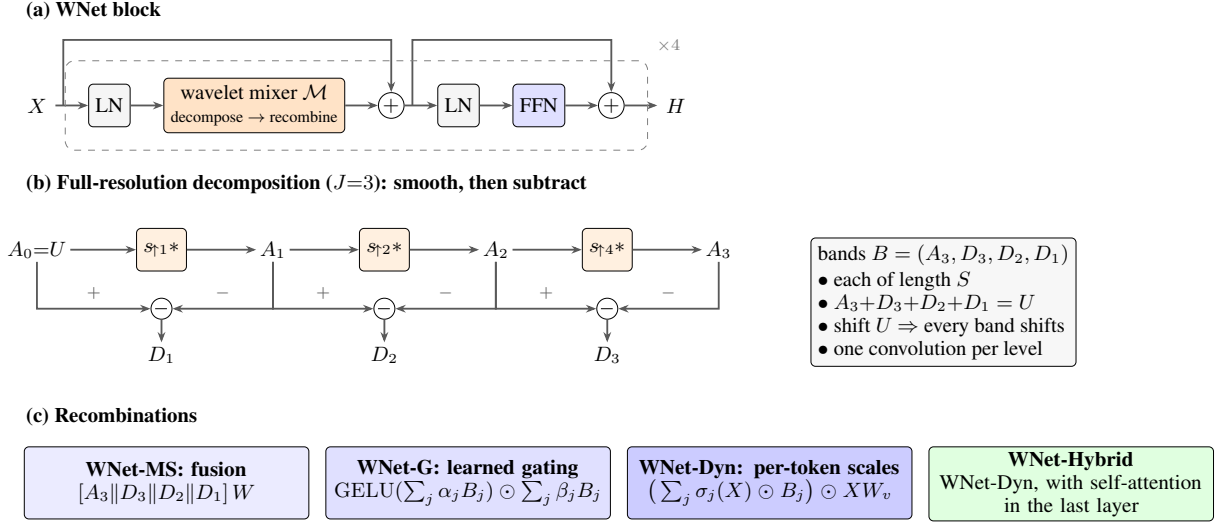

%eleborate more
WNet keeps a Transformer encoder with layer normalization before each sublayer (pre-LN) \citep{xiong2020preln} and replaces each attention sublayer with a token mixer $\mathcal{M}$:
\begin{align}
H_1 &= X + \mathcal{M}\big(\LN(X)\big),\\
H &= H_1 + \FFN\big(\LN(H_1)\big),
\end{align}
for hidden states $X\in\mathbb{R}^{S\times d}$ (Fig.~\ref{fig:arch}a). Every mixer decomposes its input along the sequence axis, separately for each channel, and then recombines the resulting bands.

\subsection{Token Mixing}
 
A WNet mixer works in two steps (Fig.~\ref{fig:arch}).
\begin{enumerate}
\setlength\itemsep{0pt}
\item \textbf{Split into scales.} Each channel of the sequence is split into $J{+}1$ bands. The \emph{smooth band} $A_J$ varies slowly along the sequence and summarises each token's wide neighbourhood. The \emph{detail bands} $D_J,\dots,D_1$ record what changes at successively finer scales, down to how a token differs from its immediate neighbours. Every band has one value per token, and we write $B_0{=}A_J$, $B_1{=}D_J,\dots,B_J{=}D_1$.

\item \textbf{Recombine.} A light learned layer combines the bands into the mixer's output. Through the smooth bands, each token draws on its context, and stacking layers widens that context.
\end{enumerate}
 
\paragraph{Implementation.} We compute the bands with the undecimated, \emph{\`a trous} DWT \citep{shensa1992wedding}. It skips the downsampling of Eq.~\ref{eq:dwt} and spreads the filter out instead (App.~\ref{app:atrous}). This has four benefits:
\begin{itemize}
\setlength\itemsep{0pt}
\item every band keeps the full length of the input (full resolution), so no interpolation is needed;
\item the bands add up exactly to the input;
\item shifting the input shifts every band by the same amount;
\item each level costs one convolution.
\end{itemize}
We explore three type of wavelet families in the context of this paper including Haar, Daubechies and Coiflets.
%use db10 with $J{=}3$ unless stated otherwise.
\subsection{Recombining the scales}

\paragraph{WNet-MS (multi-scale fusion).} The bands are concatenated and projected:
\begin{equation}
\mathcal{M}(X)=\big[A_J\,\Vert\,D_J\,\Vert\cdots\Vert\,D_1\big]\,W,
\label{eq:ms}
\end{equation}
with $W\in\mathbb{R}^{(J+1)d\times d}$. For $J{=}3$, this has the same number of parameters as attention's four projections.

\paragraph{WNet-G (learned gating).} The input is projected, $U=XW_\mathrm{i}$, and decomposed into $J{=}4$ levels. Learned per-channel weights $\alpha_j,\beta_j\in\mathbb{R}^d$ mix the bands into a gate and a signal, as in gMLP \citep{liu2021gmlp}:
\begin{equation}
\mathcal{M}(X)=\Big(\mathrm{GELU}\big(\textstyle\sum_j\alpha_j B_j\big)\odot\sum_j\beta_j B_j\Big)W_\mathrm{o}.
\end{equation}
We initialise $\alpha$ at the coarse band and $\beta$ at the details, so that coarse context starts by gating local detail.

\paragraph{WNet-Dyn (token-chosen scales).} In WNet-G, the weights that combine the bands are the same for every token. In \textbf{WNet-Dyn}, each token decides, for each group of channels, how much of each band it takes in. The mixed context then multiplies a per-token value, as in gated linear units \citep{dauphin2017glu} and Hyena \citep{poli2023hyena}:
\begin{align}
G &= \sigma(XW_g+b)\in(0,1)^{S\times(J+1)\times g},\\
\mathcal{M}(X) &= \Big(\big(\textstyle\sum_j G_j\odot B_j(XW_i)\big)\odot XW_v\Big)W_o .
\end{align}
The bands add up to their input. If all gates were equal, WNet-Dyn would therefore start out without mixing any tokens. We initialise $b$ so that each token initially reads mostly the coarse band ($\sigma(b_0){=}0.88$, $\sigma(b_{j>0}){=}0.12$). WNet-Dyn has $3d^2+(J{+}1)gd$ mixing parameters per layer, fewer than attention.

\paragraph{WNet-Hybrid.} Following FNet's hybrid \citep{leethorp2022fnet}, WNet-Hybrid is WNet-Dyn with self-attention in the last layer only.

% ===========================================================================
\section{How Far Can Information Travel?}
\label{sec:analysis}

The FFN, LayerNorm and residual connections all act on each position separately, so the mixer decides which tokens can influence each other. We measure this with the \emph{receptive field}: the number of input tokens that can influence one output position after $n$ layers. We compute it exactly by propagating each mixer's dependency pattern through the stack, including the residual path.

\paragraph{Two-tap filters confine tokens to fixed blocks.} One Haar level pairs tokens $(1,2),(3,4),\dots$. A mixer that writes each coefficient back to its own pair therefore maps every pair onto itself, and so does any stack of such layers. A token can only ever see its pair partner, however deep the network. The same argument applies to any mixer built from $J$ downsampled levels of two-tap filters: each output depends only on its aligned block of $2^J$ tokens, \emph{even if the filter weights are learned}. Learning changes how the tokens in a block are combined, not which tokens interact. In LMWT's five-level design, every token sees exactly 32 tokens after one layer and after six. A 128-token sentence is processed as four blocks that never communicate.

\begin{figure}[t]
\centering
\includegraphics[width=\columnwidth]{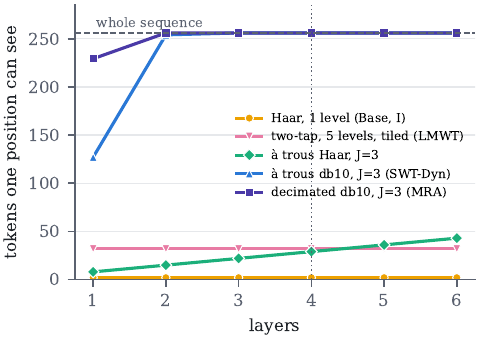}
\caption{Receptive field vs.\ depth at $S{=}256$ (exact). Two-tap designs with downsampling stay flat, whatever the depth. Full-resolution bands overlap and grow, and db10 reaches the whole sequence in two layers. Dotted line: the depth we use.}
\label{fig:reach}
\end{figure}

\paragraph{Longer filters reach the whole sequence.} A level-$j$ band of an $m$-tap filter combines $(2^j{-}1)(m{-}1){+}1$ tokens. In the full-resolution decomposition, neighbouring tokens' windows overlap, so the receptive field grows with depth even for Haar (8, 15, 22, 29 tokens over four layers; Fig.~\ref{fig:reach}). With db10, WNet-MS and WNet-Dyn see 127 tokens after one layer and the whole 256-token sequence after two. WNet-G, with four levels, sees 254 tokens after one layer.

\paragraph{Position consistency.}
\label{sec:shift}
Because the DWT of Eq.~\ref{eq:dwt} keeps every second value, its coefficients for a sequence shifted by one token are not a shifted copy of the original coefficients. A phrase is therefore represented differently depending on whether it starts at an even or an odd position. On random inputs at $S{=}256$, a one-token shift changes the three-level DWT bands by 110\% of their norm for Haar and by 67\% for db10. The full-resolution bands of Eq.~\ref{eq:atrous} are built from convolutions and differences only, so they shift exactly with the input (measured error $<10^{-6}$).

\paragraph{Cost.} The wavelet mixers cost $\mathcal{O}(JmSd)$, FNet $\mathcal{O}(Sd\log(Sd))$ \citep{cooley1965fft}, and attention $\mathcal{O}(S^2d)$ plus $\mathcal{O}(Sd^2)$ for its projections. With $d{=}256$, attention's quadratic term only overtakes its projections around $S{=}1024$, so scaling has to be measured at longer lengths (\S\ref{sec:speed}). Table~\ref{tab:mixers} summarises parameters, receptive fields and measured cost.

\begin{table*}[t]
\centering
\small
\setlength{\tabcolsep}{5pt}
\begin{tabular}{@{}llcrrrrr@{}}
\toprule
& \textbf{Mixer} & \textbf{Wavelet, $J$} & \textbf{Mixing params} & \textbf{RF$_1$} & \textbf{RF$_4$} & \textbf{ms @256} & \textbf{ms @4096} \\
\midrule
\multirow{5}{*}{\rotatebox{90}{\scriptsize baselines}}
& BERT (attention)            & --         & 1053k & 256 & 256 & 28.3 & 87.6 \\
& FNet                        & --         & 0     & 256 & 256 & 19.1 & 20.6 \\
& FNet + attn (last layer)    & --         & 263k  & 256 & 256 & 19.4 & 36.6 \\
& LMWT-style                  & Haar, 5    & 1594k & 32  & 32  & 20.0 & 20.5 \\
& No mixing (control)         & --         & 798k  & 1   & 1   & 18.6 & 19.0 \\
\midrule
\multirow{4}{*}{\rotatebox{90}{\scriptsize WNet}}
& WNet-MS                     & db10, 3    & 1050k & 127 & 256 & 27.9 & 28.1 \\
& WNet-G                      & db10, 4    & 537k  & 254 & 256 & 33.8 & 33.6 \\
& WNet-Dyn                    & db10, 3    & 823k  & 127 & 256 & 29.8 & 33.0 \\
& WNet-Hybrid                 & db10, 3    & 880k  & 127$^\ddagger$ & 256 & 28.0 & 45.6 \\
\bottomrule
\end{tabular}
\caption{Token mixers in a 4-layer encoder with $d{=}256$. Mixing parameters are summed over the four layers. RF$_n$ is the number of tokens that can influence one output position after $n$ layers at $S{=}256$ (exact). Time is one encoder training step (forward, backward and update) on one T4 at a fixed 8,192 tokens per batch, median of 20 steps; repeated runs agree within $\pm$3\%. FNet is timed with the rescaled transform (\S\ref{sec:setup}). $^\ddagger$Wavelet layers only; the attention layer is global.}
\label{tab:mixers}
\end{table*}

% ===========================================================================
\section{Experimental Setup}
\label{sec:setup}

All models share one encoder and one training pipeline; only the token mixer differs. The encoder has 4 pre-LN blocks, hidden size 256, FFN size 1024, learned position embeddings and no dropout. We pre-train with masked language modelling (15\% of tokens, 80/10/10 replacement) on a fixed subset of C4 \citep{raffel2020t5}. The text is tokenised with the FNet SentencePiece vocabulary (32k) and packed into blocks of 256 tokens without padding, presented in the same order to every model. We optimise with AdamW at $5{\times}10^{-5}$, linear decay without warm-up, batch 32, for 10,000 steps, in fp16 on one NVIDIA T4. Seeds, data order and kernel choices are fixed and recorded.

\paragraph{Baselines.}
\begin{itemize}
\setlength\itemsep{0pt}
\item \textbf{BERT} uses 8-head self-attention in the same encoder, so it is size-matched to WNet.
\item \textbf{FNet} takes the real part of a 2D Fourier transform \citep{leethorp2022fnet}. In a pre-LN encoder, the unscaled transform produces values about 180 times larger than its input and swamps the residual stream. We therefore scale it by $1/\sqrt{Sd}$, which lowers FNet's validation loss after 500 steps from 8.70 to 8.05. The unscaled results are in App.~\ref{app:glue}.
\item \textbf{FNet + attn} is FNet's own hybrid, with attention in the last layer; it is the baseline for WNet-Hybrid.
\item \textbf{LMWT-style} re-implements the mixer of \citet{kiruluta2025lmwt}: five levels of learnable two-tap filters per channel, initialised at Haar, with bands tiled to full length, concatenated and projected.
\end{itemize}

\paragraph{No-mixing control.} Gating adds multiplicative interactions that may help even without any mixing of tokens. The \textbf{no-mixing control} is therefore WNet-Dyn without any wavelet bands ($J{=}0$): it has the same gates and projections, but every token sees only itself. Whatever WNet-Dyn gains over this control comes from mixing tokens.

\paragraph{Evaluation.} We report validation MLM loss on fixed held-out blocks and masks. We also report GLUE development results \citep{wang2019glue}, fine-tuning for 3 epochs at $2{\times}10^{-5}$ with mean pooling. The models in Table~\ref{tab:main} are pre-trained with three seeds and the ablations with one seed and three fine-tuning seeds. We test differences to BERT with Welch's $t$-test. At this budget, CoLA, MRPC, RTE and STS-B stay near the majority-class rate for every model, BERT included. We report them in App.~\ref{app:glue} and use SST-2, QNLI and MNLI as the core tasks. Speed is the median training-step time of the encoder at a fixed 8,192 tokens per batch.

% ===========================================================================
\section{Results}
\label{sec:results}

\subsection{Efficiency}
\label{sec:speed}

\begin{figure}[t]
\centering
\includegraphics[width=\columnwidth]{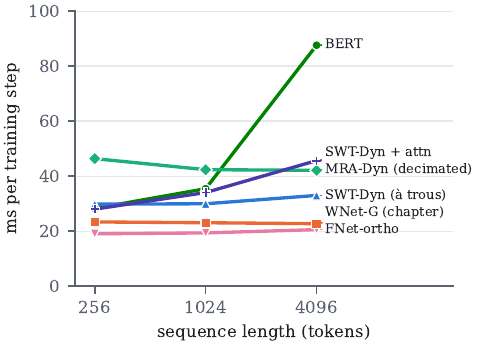}
\caption{Training-step time at a fixed 8,192 tokens per batch on one T4. Attention slows down with length; the attention-free WNet mixers do not. 
%\todo{Add $S\in\{512,2048,8192\}$ from the full run.}
}
\label{fig:speed}
\end{figure}

At a fixed number of tokens per batch, BERT's step time grows from 28 to 88\,ms between 256 and 4,096 tokens, whereas the attention-free WNet mixers stay flat (Fig.~\ref{fig:speed}, Table~\ref{tab:mixers}).
\begin{itemize}
\setlength\itemsep{0pt}
\item WNet-Dyn costs the same as BERT at 256 tokens (29.8 vs.\ 28.3\,ms) and is 2.7$\times$ faster at 4,096.
\item WNet-MS is the fastest WNet mixer: 3.1$\times$ faster than BERT at 4,096 tokens.
\item WNet-Hybrid inherits part of attention's growth from its single attention layer, but is still 1.9$\times$ faster than BERT at 4,096 tokens.
\item FNet remains the fastest model overall, at about 19\,ms.
\end{itemize}
The full-resolution decomposition is also faster than the downsampled DWT followed by its inverse: 22\% at 4,096 tokens for WNet-Dyn (Table~\ref{tab:ablation}).

\subsection{Pre-training and GLUE}

\begin{table}[t]
\centering
\footnotesize
\setlength{\tabcolsep}{1.6pt}
\begin{tabular}{@{}lccccc@{}}
\toprule
\textbf{Model} & \textbf{Loss}$\downarrow$ & \textbf{SST-2} & \textbf{QNLI} & \textbf{MNLI m/mm} & \textbf{Core} \\
\midrule
Majority      & --    & 50.9 & 50.5 & 35.4/35.2 & 45.6 \\
BERT          & 7.946 & 62.3{\tiny$\pm$0.7} & 56.4{\tiny$\pm$0.3} & 34.1/34.2 & 50.9 \\
FNet          & 8.048 & 50.9{\tiny$\pm$0.5} & 54.9{\tiny$\pm$0.6} & 34.0/34.3 & 46.6 \\
FNet + attn   & 8.606 & 54.8{\tiny$\pm$0.4} & 54.8{\tiny$\pm$0.1} & 33.6/33.8 & 47.7 \\
LMWT-style    & 7.905 & 51.1{\tiny$\pm$0.2} & 55.2{\tiny$\pm$0.4} & 33.4/33.5 & 46.6 \\
No mixing     & 7.921 & 61.0{\tiny$\pm$0.3} & 57.3{\tiny$\pm$0.2} & 34.9/34.7 & 51.1 \\
\midrule
WNet-MS       & 7.986 & 54.8{\tiny$\pm$1.3} & 56.6{\tiny$\pm$0.3} & 34.9/35.5 & 48.8 \\
WNet-G        & 7.925 & 55.7{\tiny$\pm$2.2} & 56.3{\tiny$\pm$0.6} & 34.8/34.7 & 48.9 \\
WNet-Dyn      & 7.919 & 56.0{\tiny$\pm$0.1} & 57.9{\tiny$\pm$0.2} & 34.5/34.4 & 49.5 \\
WNet-Hybrid   & 7.902 & 58.1{\tiny$\pm$3.3} & 57.8{\tiny$\pm$0.1} & 34.5/35.2 & 50.1 \\
\bottomrule
\end{tabular}
\caption{Validation MLM loss and GLUE development accuracy on the core tasks (subscript: s.d.); Core averages SST-2, QNLI and MNLI-m. These numbers come from a quick run that validates the pipeline: 500 pre-training steps (one seed), with GLUE fine-tuned on 5,000 examples per task, mean of two fine-tuning seeds. 
%\todo{Replace with the 10k-step, three-seed run.}
}
\label{tab:main}
\end{table}

Table~\ref{tab:main} reports the quick run (500 steps).
\begin{itemize}
\setlength\itemsep{0pt}
\item \textbf{Loss.} WNet-Hybrid has the lowest loss of all models (7.902, against 7.946 for BERT). WNet-Dyn (7.919) and WNet-G (7.925) are also below BERT; WNet-MS is above it (7.986).
\item \textbf{The control is already competitive.} The no-mixing control reaches 7.921, as low as WNet-Dyn. At this stage the models are still learning word frequencies and very local statistics, so mixing tokens does not yet help (\S\ref{sec:discussion}).
\item \textbf{GLUE.} MNLI is at chance for every model (33.4--35.5 against a majority rate of 35.4). On SST-2, BERT scores highest (62.3), followed by the control (61.0) and WNet-Hybrid (58.1). On QNLI, WNet-Dyn (57.9) and WNet-Hybrid (57.8) score above BERT (56.4). Every WNet mixer is ahead of FNet and LMWT-style on the core average (48.8--50.1 vs.\ 46.6).
\end{itemize}

A preliminary run with the full protocol (10,000 steps, three fine-tuning seeds) shows which task separates the mixers. A mixer with a single Haar level, in which every token sees only its pair partner, matched BERT on SST-2 (80.2 vs.\ 79.7) and nearly matched it on QNLI (61.0 vs.\ 61.4). It trailed BERT by 4.8 points on MNLI-m (49.6 vs.\ 54.4), and FNet trailed by 7.6 (46.8). With mean pooling, sentiment and question--answer matching at this scale can largely be solved from word identity. MNLI, by contrast, requires relating the premise to the hypothesis. It is therefore the test of whether a mixer relates tokens usefully, and a pair-local mixer falls short on it, as the analysis predicts. 
%\todo{Full run: do WNet-Dyn and WNet-Hybrid close the MNLI gap, and do they beat the no-mixing control and LMWT-style?}

\subsection{Ablations}

\begin{table}[t]
\centering
\small
\setlength{\tabcolsep}{2pt}
\begin{tabular}{@{}lccc@{}}
\toprule
\textbf{Variant} & \textbf{Loss} & \textbf{MNLI-m} & \textbf{ms@4k} \\
\midrule
WNet-Dyn (db10, $J{=}3$)                  & 7.919 & 34.5 & 33.0 \\
\quad downsampled DWT + inverse           & 7.920 & 34.5 & 42.1 \\
\quad Haar instead of db10                & 7.920 & 34.4 & 30.8 \\
\quad learned filter (init.\ db10)        & 7.916 & 34.6 & 35.1 \\
\quad no wavelet bands ($J{=}0$)          & 7.921 & 34.9 & 19.0 \\
\midrule
WNet-MS (db10, $J{=}3$)                   & 7.986 & 34.9 & 28.1 \\
\quad downsampled DWT + inverse           & 7.948 & 35.2 & 38.3 \\
\quad downsampled + interpolation$^\ast$  & 7.951 & 33.3 & 30.1 \\
\midrule
WNet-G (db10, $J{=}4$)                    & 7.925 & 34.8 & 33.6 \\
\quad downsampled DWT + inverse           & 7.928 & 35.2 & 54.2 \\
\bottomrule
\end{tabular}
\caption{Ablations, changing one component at a time. ``Downsampled DWT + inverse'' computes Eq.~\ref{eq:dwt} and returns each band to full length with the inverse DWT; ``+ interpolation'' stretches the bands by linear interpolation instead ($^\ast$coif4). ms@4k: step time at 4,096 tokens. Quick run as in Table~\ref{tab:main}. %\todo{Replace with the 10k-step run.}
}
\label{tab:ablation}
\end{table}

Table~\ref{tab:ablation} varies one component at a time, the choice of transform, wavelet and filter barely changes the loss, and removing mixing altogether costs only 0.002. The efficiency differences, by contrast, are final. The full-resolution decomposition is 22\% faster than the downsampled DWT with its inverse for WNet-Dyn, 27\% for WNet-MS and 38\% for WNet-G, because each band is one convolution and nothing has to be reconstructed. Haar is cheaper than db10. 
%\todo{Full run: report which components matter for loss and MNLI.}

% ===========================================================================
\section{Discussion}
\label{sec:discussion}

\paragraph{Early loss rewards locality, not mixing.} After 500 steps, the no-mixing control matches the wavelet mixers (7.921 vs.\ 7.919--7.920), and both beat BERT (7.946). A short wavelet sweep also favours the shortest filter (App.~\ref{app:early}). Short runs therefore cannot evaluate token mixing, and choosing a wavelet by early loss would systematically favour short filters. We therefore chose db10 in advance, for its receptive field.

\paragraph{The receptive field can be checked before training.} A pair- or block-local mixer cannot relate distant tokens, however long it is trained. The check takes no training and a few lines of code. It explains the MNLI gap of the single-level Haar mixer, and it applies to published designs such as learnable Haar mixing. %\todo{Connect to the LMWT-style result of the full run.}

\paragraph{Fixed mixing and content.} A fixed mixer such as WNet-MS combines two tokens in the same way whatever the words are. That suits tasks decided by local cues, but not tasks that require aligning two sentences. WNet-Dyn keeps the wavelet transform fixed and linear-time, but lets each token choose its scales; WNet-Hybrid adds one layer of full content-based attention. 
%\todo{Interpret the WNet-Dyn and WNet-Hybrid results.}

% ===========================================================================
\section{Conclusion}

WNet replaces self-attention with wavelet token mixing. It decomposes the hidden states into scales that keep the full length of the sequence, add up to the input and shift with it, and then recombines them with a light learned layer. How far information can travel depends on the filter: two-tap designs, learned or fixed, are confined to fixed blocks, whereas longer filters reach the whole sequence within two layers. WNet-Dyn trains as fast as attention at 256 tokens and 2.7$\times$ faster at 4,096. 
%\todo{Final sentence on quality relative to BERT and the no-mixing control.}

\section*{Limitations}

\textbf{Scale.} All models have four layers and hidden size 256 and are pre-trained for 10,000 steps on one T4. Absolute GLUE scores are far below published BERT and FNet results, four GLUE tasks stay near the majority class, and rankings may change at larger scale. \textbf{Statistics.} We use three runs per model for the comparisons that carry claims, so only large, consistent differences reach significance. \textbf{Length.} We measure cost up to 4,096 tokens but quality only at 256, and make no claim about long-context accuracy. \textbf{Determinism.} Some GPU kernels (attention's backward pass, reflection padding) are not deterministic, so reruns of those models agree to about three decimals. \textbf{Scope.} We study English bidirectional encoders. Causal decoding would need one-sided filters, since a centred filter lets a token see the next one.

\section*{Ethical Considerations}
We pre-train on C4, a filtered web corpus that contains social biases and may contain personal information \citep{raffel2020t5}. Our models are small research artefacts not intended for deployment. All experiments ran on a single GPU.

\bibliography{refs}

@inproceedings{vaswani2017attention,
  title     = {Attention is All you Need},
  author    = {Vaswani, Ashish and Shazeer, Noam and Parmar, Niki and Uszkoreit, Jakob and Jones, Llion and Gomez, Aidan N. and Kaiser, {\L}ukasz and Polosukhin, Illia},
  booktitle = {Advances in Neural Information Processing Systems},
  volume    = {30},
  year      = {2017}
}

@inproceedings{devlin2019bert,
  title     = {BERT: Pre-training of Deep Bidirectional Transformers for Language Understanding},
  author    = {Devlin, Jacob and Chang, Ming-Wei and Lee, Kenton and Toutanova, Kristina},
  booktitle = {Proceedings of the 2019 Conference of the North {A}merican Chapter of the Association for Computational Linguistics: Human Language Technologies, Volume 1 (Long and Short Papers)},
  pages     = {4171--4186},
  year      = {2019}
}

@inproceedings{leethorp2022fnet,
  title     = {FNet: Mixing Tokens with {F}ourier Transforms},
  author    = {Lee-Thorp, James and Ainslie, Joshua and Eckstein, Ilya and Onta{\~n}{\'o}n, Santiago},
  booktitle = {Proceedings of the 2022 Conference of the North American Chapter of the Association for Computational Linguistics: Human Language Technologies},
  pages     = {4296--4313},
  year      = {2022}
}

@article{raffel2020t5,
  title   = {Exploring the Limits of Transfer Learning with a Unified Text-to-Text Transformer},
  author  = {Raffel, Colin and Shazeer, Noam and Roberts, Adam and Lee, Katherine and Narang, Sharan and Matena, Michael and Zhou, Yanqi and Li, Wei and Liu, Peter J.},
  journal = {Journal of Machine Learning Research},
  volume  = {21},
  number  = {140},
  pages   = {1--67},
  year    = {2020}
}

@inproceedings{wang2019glue,
  title     = {GLUE: A Multi-Task Benchmark and Analysis Platform for Natural Language Understanding},
  author    = {Wang, Alex and Singh, Amanpreet and Michael, Julian and Hill, Felix and Levy, Omer and Bowman, Samuel R.},
  booktitle = {International Conference on Learning Representations},
  year      = {2019}
}

@article{mallat1989theory,
  title   = {A Theory for Multiresolution Signal Decomposition: The Wavelet Representation},
  author  = {Mallat, St{\'e}phane G.},
  journal = {IEEE Transactions on Pattern Analysis and Machine Intelligence},
  volume  = {11},
  number  = {7},
  pages   = {674--693},
  year    = {1989}
}

@book{daubechies1992ten,
  title     = {Ten Lectures on Wavelets},
  author    = {Daubechies, Ingrid},
  publisher = {Society for Industrial and Applied Mathematics},
  year      = {1992}
}

@article{cooley1965fft,
  title   = {An Algorithm for the Machine Calculation of Complex {F}ourier Series},
  author  = {Cooley, James W. and Tukey, John W.},
  journal = {Mathematics of Computation},
  volume  = {19},
  number  = {90},
  pages   = {297--301},
  year    = {1965}
}

@misc{liu2021gmlp,
      title={Pay Attention to MLPs}, 
      author={Hanxiao Liu and Zihang Dai and David R. So and Quoc V. Le},
      year={2021},
      eprint={2105.08050},
      archivePrefix={arXiv},
      primaryClass={cs.LG},
      url={https://arxiv.org/abs/2105.08050}, 
}

@misc{tolstikhin2021mlpmixer,
      title={MLP-Mixer: An all-MLP Architecture for Vision}, 
      author={Ilya Tolstikhin and Neil Houlsby and Alexander Kolesnikov and Lucas Beyer and Xiaohua Zhai and Thomas Unterthiner and Jessica Yung and Andreas Steiner and Daniel Keysers and Jakob Uszkoreit and Mario Lucic and Alexey Dosovitskiy},
      year={2021},
      eprint={2105.01601},
      archivePrefix={arXiv},
      primaryClass={cs.CV},
      url={https://arxiv.org/abs/2105.01601}, 
}

@inproceedings{rao2021gfnet,
  title     = {Global Filter Networks for Image Classification},
  author    = {Rao, Yongming and Zhao, Wenliang and Zhu, Zheng and Lu, Jiwen and Zhou, Jie},
  booktitle = {Advances in Neural Information Processing Systems},
  volume    = {34},
  pages     = {980--993},
  year      = {2021}
}

@inproceedings{guibas2022afno,
  title     = {Efficient Token Mixing for Transformers via Adaptive {F}ourier Neural Operators},
  author    = {Guibas, John and Mardani, Morteza and Li, Zongyi and Tao, Andrew and Anandkumar, Anima and Catanzaro, Bryan},
  booktitle = {International Conference on Learning Representations},
  year      = {2022}
}

@misc{jeevan2022wavemix,
      title={WaveMix: Resource-efficient Token Mixing for Images}, 
      author={Pranav Jeevan and Amit Sethi},
      year={2022},
      eprint={2203.03689},
      archivePrefix={arXiv},
      primaryClass={cs.CV},
      url={https://arxiv.org/abs/2203.03689}, 
}

@misc{yao2022wavevit,
      title={Wave-ViT: Unifying Wavelet and Transformers for Visual Representation Learning}, 
      author={Ting Yao and Yingwei Pan and Yehao Li and Chong-Wah Ngo and Tao Mei},
      year={2022},
      eprint={2207.04978},
      archivePrefix={arXiv},
      primaryClass={cs.CV},
      url={https://arxiv.org/abs/2207.04978}, 
}

@misc{patro2023svt,
      title={Scattering Vision Transformer: Spectral Mixing Matters}, 
      author={Badri N. Patro and Vijay Srinivas Agneeswaran},
      year={2023},
      eprint={2311.01310},
      archivePrefix={arXiv},
      primaryClass={cs.CV},
      url={https://arxiv.org/abs/2311.01310}, 
}

@misc{zhuang2024wavspa,
      title={WavSpA: Wavelet Space Attention for Boosting Transformers' Long Sequence Learning Ability}, 
      author={Yufan Zhuang and Zihan Wang and Fangbo Tao and Jingbo Shang},
      year={2023},
      eprint={2210.01989},
      archivePrefix={arXiv},
      primaryClass={cs.CL},
      url={https://arxiv.org/abs/2210.01989}, 
}

@misc{kiruluta2025lmwt,
      title={Learnable Multi-Scale Wavelet Transformer: A Novel Alternative to Self-Attention}, 
      author={Andrew Kiruluta and Priscilla Burity and Samantha Williams},
      year={2025},
      eprint={2504.08801},
      archivePrefix={arXiv},
      primaryClass={cs.LG},
      url={https://arxiv.org/abs/2504.08801}, 
}

@inproceedings{he2023fouriertransformer,
   title={Fourier Transformer: Fast Long Range Modeling by Removing Sequence Redundancy with FFT Operator},
   url={http://dx.doi.org/10.18653/v1/2023.findings-acl.570},
   DOI={10.18653/v1/2023.findings-acl.570},
   booktitle={Findings of the Association for Computational Linguistics: ACL 2023},
   publisher={Association for Computational Linguistics},
   author={He, Ziwei and Yang, Meng and Feng, Minwei and Yin, Jingcheng and Wang, Xinbing and Leng, Jingwen and Lin, Zhouhan},
   year={2023},
   pages={8954–8966} }

@article{scribano2023dctformer,
   title={DCT-Former: Efficient Self-Attention with Discrete Cosine Transform},
   volume={94},
   ISSN={1573-7691},
   url={http://dx.doi.org/10.1007/s10915-023-02125-5},
   DOI={10.1007/s10915-023-02125-5},
   number={3},
   journal={Journal of Scientific Computing},
   publisher={Springer Science and Business Media LLC},
   author={Scribano, Carmelo and Franchini, Giorgia and Prato, Marco and Bertogna, Marko},
   year={2023},
   month=Feb }

@misc{dai2020funnel,
      title={Funnel-Transformer: Filtering out Sequential Redundancy for Efficient Language Processing}, 
      author={Zihang Dai and Guokun Lai and Yiming Yang and Quoc V. Le},
      year={2020},
      eprint={2006.03236},
      archivePrefix={arXiv},
      primaryClass={cs.LG},
      url={https://arxiv.org/abs/2006.03236}, 
}

@misc{poli2023hyena,
      title={Hyena Hierarchy: Towards Larger Convolutional Language Models}, 
      author={Michael Poli and Stefano Massaroli and Eric Nguyen and Daniel Y. Fu and Tri Dao and Stephen Baccus and Yoshua Bengio and Stefano Ermon and Christopher Ré},
      year={2023},
      eprint={2302.10866},
      archivePrefix={arXiv},
      primaryClass={cs.LG},
      url={https://arxiv.org/abs/2302.10866}, 
}

@misc{fu2023monarch,
      title={Monarch Mixer: A Simple Sub-Quadratic GEMM-Based Architecture}, 
      author={Daniel Y. Fu and Simran Arora and Jessica Grogan and Isys Johnson and Sabri Eyuboglu and Armin W. Thomas and Benjamin Spector and Michael Poli and Atri Rudra and Christopher Ré},
      year={2023},
      eprint={2310.12109},
      archivePrefix={arXiv},
      primaryClass={cs.LG},
      url={https://arxiv.org/abs/2310.12109}, 
}

@misc{gu2022s4,
      title={Efficiently Modeling Long Sequences with Structured State Spaces}, 
      author={Albert Gu and Karan Goel and Christopher Ré},
      year={2022},
      eprint={2111.00396},
      archivePrefix={arXiv},
      primaryClass={cs.LG},
      url={https://arxiv.org/abs/2111.00396}, 
}

@misc{gu2024mamba,
      title={Mamba: Linear-Time Sequence Modeling with Selective State Spaces}, 
      author={Albert Gu and Tri Dao},
      year={2024},
      eprint={2312.00752},
      archivePrefix={arXiv},
      primaryClass={cs.LG},
      url={https://arxiv.org/abs/2312.00752}, 
}

@misc{feinashley2025spectre,
      title={SPECTRE: An FFT-Based Efficient Drop-In Replacement to Self-Attention for Long Contexts}, 
      author={Jacob Fein-Ashley and Neelesh Gupta and Rajgopal Kannan and Viktor Prasanna},
      year={2025},
      eprint={2502.18394},
      archivePrefix={arXiv},
      primaryClass={cs.LG},
      url={https://arxiv.org/abs/2502.18394}, 
}

@misc{gan2026caracal,
      title={Caracal: Causal Architecture via Spectral Mixing}, 
      author={Bingzheng Gan and Tianyi Zhang and Yusu Li and Jing Huang and Wei Shi and Yangkai Ding and Tao Yu},
      year={2026},
      eprint={2605.00292},
      archivePrefix={arXiv},
      primaryClass={cs.LG},
      url={https://arxiv.org/abs/2605.00292}, 
}

@misc{tay2022survey,
      title={Efficient Transformers: A Survey}, 
      author={Yi Tay and Mostafa Dehghani and Dara Bahri and Donald Metzler},
      year={2022},
      eprint={2009.06732},
      archivePrefix={arXiv},
      primaryClass={cs.LG},
      url={https://arxiv.org/abs/2009.06732}, 
}

@misc{wang2020linformer,
      title={Linformer: Self-Attention with Linear Complexity}, 
      author={Sinong Wang and Belinda Z. Li and Madian Khabsa and Han Fang and Hao Ma},
      year={2020},
      eprint={2006.04768},
      archivePrefix={arXiv},
      primaryClass={cs.LG},
      url={https://arxiv.org/abs/2006.04768}, 
}

@misc{choromanski2021performer,
      title={Rethinking Attention with Performers}, 
      author={Krzysztof Choromanski and Valerii Likhosherstov and David Dohan and Xingyou Song and Andreea Gane and Tamas Sarlos and Peter Hawkins and Jared Davis and Afroz Mohiuddin and Lukasz Kaiser and David Belanger and Lucy Colwell and Adrian Weller},
      year={2022},
      eprint={2009.14794},
      archivePrefix={arXiv},
      primaryClass={cs.LG},
      url={https://arxiv.org/abs/2009.14794}, 
}

@article{beltagy2020longformer,
  author       = {Iz Beltagy and
                  Matthew E. Peters and
                  Arman Cohan},
  title        = {Longformer: The Long-Document Transformer},
  journal      = {CoRR},
  volume       = {abs/2004.05150},
  year         = {2020},
  url          = {https://doi.org/10.48550/arXiv.2004.05150},
  doi          = {10.48550/ARXIV.2004.05150},
  eprinttype   = {arXiv},
  eprint       = {2004.05150},
  bibsource    = {dblp computer science bibliography, https://dblp.org}
}

@article{zaheer2020bigbird,
  author       = {Iz Beltagy and
                  Matthew E. Peters and
                  Arman Cohan},
  title        = {Longformer: The Long-Document Transformer},
  journal      = {CoRR},
  volume       = {abs/2004.05150},
  year         = {2020},
  url          = {https://doi.org/10.48550/arXiv.2004.05150},
  doi          = {10.48550/ARXIV.2004.05150},
  eprinttype   = {arXiv},
  eprint       = {2004.05150},
  bibsource    = {dblp computer science bibliography, https://dblp.org}
}

@misc{xiong2020preln,
      title={On Layer Normalization in the Transformer Architecture}, 
      author={Ruibin Xiong and Yunchang Yang and Di He and Kai Zheng and Shuxin Zheng and Chen Xing and Huishuai Zhang and Yanyan Lan and Liwei Wang and Tie-Yan Liu},
      year={2020},
      eprint={2002.04745},
      archivePrefix={arXiv},
      primaryClass={cs.LG},
      url={https://arxiv.org/abs/2002.04745}, 
}

@misc{dauphin2017glu,
      title={Language Modeling with Gated Convolutional Networks}, 
      author={Yann N. Dauphin and Angela Fan and Michael Auli and David Grangier},
      year={2017},
      eprint={1612.08083},
      archivePrefix={arXiv},
      primaryClass={cs.CL},
      url={https://arxiv.org/abs/1612.08083}, 
}

@misc{verma2024waveletgpt,
      title={Wavelet GPT: Wavelet Inspired Large Language Models}, 
      author={Prateek Verma},
      year={2025},
      eprint={2409.12924},
      archivePrefix={arXiv},
      primaryClass={eess.SP},
      url={https://arxiv.org/abs/2409.12924}, 
}

@ARTICLE{shensa1992wedding,
  author={Shensa, M.J.},
  journal={IEEE Transactions on Signal Processing}, 
  title={The discrete wavelet transform: wedding the a trous and Mallat algorithms}, 
  year={1992},
  volume={40},
  number={10},
  pages={2464-2482},
  doi={10.1109/78.157290}}

% ===========================================================================
\appendix

\section{The DWT in One Picture}
\label{app:dwt}

\begin{figure}[H]
\centering
\includegraphics[width=\columnwidth]{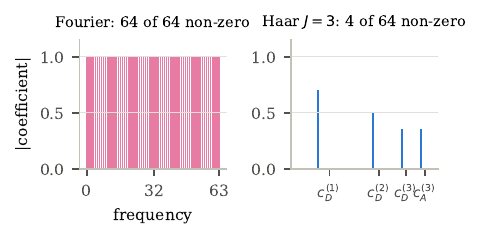}
\caption{Coefficients affected by changing one token ($S{=}64$, position 21). The Fourier transform spreads the change over all coefficients with equal magnitude; a three-level Haar DWT confines it to one coefficient per band.}
\label{fig:impulse}
\end{figure}

Figure~\ref{fig:impulse} contrasts the Fourier transform with the DWT. Changing one token changes every Fourier coefficient by the same amount. In a DWT, it changes only the few coefficients whose region contains that token. WNet's full-resolution decomposition keeps every band at the length of the input. It does so by spreading the filter out at each level instead of keeping every second value.

\section{The \`A Trous Decomposition}
\label{app:atrous}
 
Let $s$ be the wavelet's low-pass filter normalised to unit sum, and $s_{\up k}$ the same filter with $k{-}1$ zeros between its taps. Starting from the mixer's input, $A_0=U$, each level smooths the previous approximation and keeps what the smoothing removed as a detail:
\begin{equation}
A_j=s_{\up 2^{j-1}}*A_{j-1},\qquad D_j=A_{j-1}-A_j,
\label{eq:atrous}
\end{equation}
with circular convolution (Fig.~\ref{fig:arch}b). The bands add up to the input by telescoping, $A_J+\sum_{j}D_j=A_0$, for any filter $s$, including a learned one. Because Eq.~\ref{eq:atrous} contains only convolutions and differences, which commute with shifts, a shifted input yields shifted bands. Level $j$ is one dilated depthwise convolution, and no inverse transform is needed.

\section{Implementation Details}
\label{app:hparams}

\begin{table}[H]
\centering
\footnotesize
\setlength{\tabcolsep}{2.5pt}
\begin{tabular}{@{}ll@{}}
\toprule
Layers / hidden / FFN      & 4 / 256 / 1024, pre-LN \\
Attention heads            & 8 (BERT, hybrids) \\
Tokenizer                  & FNet SentencePiece (32{,}000) \\
Sequence length            & 256, packed (no padding) \\
Pre-training data          & C4, 85M tokens (one pass) \\
Objective                  & MLM 15\%, 80/10/10 \\
Batch / steps              & 32 / 10{,}000 \\
Optimizer                  & AdamW (0.9, 0.999), no wd \\
LR / schedule              & $5{\times}10^{-5}$, linear decay, no warm-up \\
Clipping / dropout         & 1.0 / 0 \\
Wavelet                    & db10 (Haar, coif4 in ablations) \\
Levels $J$                 & 3 (MS, Dyn, Hybrid); 4 (G) \\
Levels $J$ (LMWT-style)    & 5 \\
Gate groups (Dyn)          & 8 \\
GLUE                       & 3 ep., lr $2{\times}10^{-5}$, bs 32, mean pool \\
Pre-training seeds         & 3 (Table~\ref{tab:main}) \\
Hardware / precision       & 1$\times$ T4 (16\,GB) / fp16 \\
\bottomrule
\end{tabular}
\caption{Hyperparameters shared by all models.}
\end{table}

Every transform is checked on the GPU before training. The downsampled DWT is compared against \texttt{pytorch\_wavelets} (maximum error 0.0) and interpolation against \texttt{F.interpolate} ($2.4{\times}10^{-7}$). The full-resolution decomposition is checked for exact reconstruction and shift behaviour ($<10^{-6}$ and $0.0$). The vocabulary head is evaluated only on masked positions, which leaves the loss unchanged. Equivalent implementations of the same operation are timed once and the fastest is reused by every run. All models are compiled with \texttt{torch.compile}.

\section{Full Results}
\label{app:glue}

\begin{table}[H]
\centering
\scriptsize
\setlength{\tabcolsep}{2pt}
\begin{tabular}{@{}lccccc@{}}
\toprule
\textbf{Model} & \textbf{Loss} & \textbf{SST-2} & \textbf{QNLI} & \textbf{MNLI-m} & \textbf{MNLI-mm} \\
\midrule
Majority class & -- & 50.9 & 50.5 & 35.4 & 35.2 \\
\midrule
BERT                    & 7.946 & 62.3{\tiny$\pm$0.7} & 56.4{\tiny$\pm$0.3} & 34.1{\tiny$\pm$0.6} & 34.2{\tiny$\pm$0.2} \\
FNet (unscaled)         & 8.698 & 53.5{\tiny$\pm$1.1} & 54.5{\tiny$\pm$0.1} & 33.8{\tiny$\pm$0.3} & 34.0{\tiny$\pm$0.1} \\
FNet (scaled)           & 8.048 & 50.9{\tiny$\pm$0.5} & 54.9{\tiny$\pm$0.6} & 34.0{\tiny$\pm$0.8} & 34.3{\tiny$\pm$1.4} \\
FNet + attn             & 8.606 & 54.8{\tiny$\pm$0.4} & 54.8{\tiny$\pm$0.1} & 33.6{\tiny$\pm$0.6} & 33.8{\tiny$\pm$0.1} \\
LMWT-style              & 7.905 & 51.1{\tiny$\pm$0.2} & 55.2{\tiny$\pm$0.4} & 33.4{\tiny$\pm$0.2} & 33.5{\tiny$\pm$0.1} \\
No mixing               & 7.921 & 61.0{\tiny$\pm$0.3} & 57.3{\tiny$\pm$0.2} & 34.9{\tiny$\pm$0.1} & 34.7{\tiny$\pm$0.2} \\
\midrule
WNet-MS                 & 7.986 & 54.8{\tiny$\pm$1.3} & 56.6{\tiny$\pm$0.3} & 34.9{\tiny$\pm$0.0} & 35.5{\tiny$\pm$0.1} \\
WNet-G                  & 7.925 & 55.7{\tiny$\pm$2.2} & 56.3{\tiny$\pm$0.6} & 34.8{\tiny$\pm$1.2} & 34.7{\tiny$\pm$0.6} \\
WNet-Dyn                & 7.919 & 56.0{\tiny$\pm$0.1} & 57.9{\tiny$\pm$0.2} & 34.5{\tiny$\pm$0.4} & 34.4{\tiny$\pm$0.1} \\
WNet-Hybrid             & 7.902 & 58.1{\tiny$\pm$3.3} & 57.8{\tiny$\pm$0.1} & 34.5{\tiny$\pm$0.3} & 35.2{\tiny$\pm$0.1} \\
\midrule
\multicolumn{6}{@{}l}{\textit{Ablations}} \\
MS, DWT + inverse       & 7.948 & 55.7{\tiny$\pm$1.3} & 58.0{\tiny$\pm$0.1} & 35.2{\tiny$\pm$0.3} & 34.6{\tiny$\pm$0.1} \\
MS, interpolation$^\ast$& 7.951 & 53.0{\tiny$\pm$0.2} & 54.2{\tiny$\pm$0.2} & 33.3{\tiny$\pm$0.2} & 33.6{\tiny$\pm$0.0} \\
G, DWT + inverse        & 7.928 & 59.7{\tiny$\pm$2.2} & 58.0{\tiny$\pm$0.2} & 35.2{\tiny$\pm$0.4} & 34.7{\tiny$\pm$0.1} \\
Dyn, DWT + inverse      & 7.920 & 56.8{\tiny$\pm$0.2} & 57.5{\tiny$\pm$0.0} & 34.5{\tiny$\pm$0.5} & 34.5{\tiny$\pm$0.1} \\
Dyn, learned filter     & 7.916 & 56.4{\tiny$\pm$0.9} & 58.0{\tiny$\pm$0.0} & 34.6{\tiny$\pm$0.3} & 34.3{\tiny$\pm$0.2} \\
Dyn, Haar               & 7.920 & 56.7{\tiny$\pm$0.1} & 57.4{\tiny$\pm$0.2} & 34.4{\tiny$\pm$0.4} & 34.5{\tiny$\pm$0.1} \\
\bottomrule
\end{tabular}
\caption{All models in the quick run (500 pre-training steps, one seed; GLUE fine-tuned on 5,000 examples per task, mean of two fine-tuning seeds). ``DWT + inverse'': downsampled bands returned to full length with the inverse DWT; $^\ast$coif4. The quick run evaluates only the core tasks. 
%\todo{Replace with the full run, including CoLA, MRPC, RTE and STS-B.}
}
\end{table}

\section{Early Training and the Wavelet Sweep}
\label{app:early}

\begin{table}[H]
\centering
\footnotesize
\setlength{\tabcolsep}{3pt}
\begin{tabular}{@{}lc@{\hspace{8pt}}lc@{}}
\toprule
\multicolumn{2}{c}{\textbf{Models, 500 steps}} & \multicolumn{2}{c}{\textbf{WNet-MS, 300 steps}} \\
\midrule
LMWT-style          & 7.905 & Haar  & 8.614 \\
WNet-Dyn (db10)     & 7.919 & sym10 & 8.628 \\
WNet-Dyn (Haar)     & 7.920 & db4   & 8.660 \\
No mixing           & 7.921 & coif4 & 8.663 \\
BERT                & 7.946 & db20  & 8.667 \\
FNet (scaled)       & 8.048 & db10  & 8.668 \\
FNet (unscaled)     & 8.698 &       &       \\
\bottomrule
\end{tabular}
\caption{Validation loss early in training (one seed). Left: a model that cannot mix tokens matches the wavelet mixers and beats attention. Right: short runs favour short filters. Neither ordering reflects the ability to relate distant tokens.}
\end{table}

\end{document}